\documentclass[10pt, a4paper]{article}
\usepackage{lrec2022} 
\usepackage{multibib}
\newcites{languageresource}{Language Resources}
\usepackage{graphicx}
\usepackage{tabularx}
\usepackage{soul}
\usepackage{titlesec}
\titleformat{\section}{\normalfont\large\bfseries\center}{\thesection.}{1em}{}
\titleformat{\subsection}{\normalfont\SmallTitleFont\bfseries\raggedright}{\thesubsection.}{1em}{}
\titleformat{\subsubsection}{\normalfont\normalsize\bfseries\raggedright}{\thesubsubsection.}{1em}{}
\renewcommand\thesection{\arabic{section}}
\renewcommand\thesubsection{\thesection.\arabic{subsection}}
\renewcommand\thesubsubsection{\thesubsection.\arabic{subsubsection}}

\usepackage{epstopdf}
\usepackage[utf8]{inputenc}

\usepackage{hyperref}
\usepackage{xstring}

\usepackage{color}

\usepackage{booktabs}

\title{NLU for Game-based Learning in Real: Initial Evaluations}

\name{Eda Okur, Saurav Sahay, Lama Nachman} 

\address{Intel Labs \\
         USA \\
         \{eda.okur, saurav.sahay, lama.nachman\}@intel.com\\}

\abstract{
Intelligent systems designed for play-based interactions should be contextually aware of the users and their surroundings. Spoken Dialogue Systems (SDS) are critical for these interactive agents to carry out effective goal-oriented communication with users in real-time. For the real-world (i.e., in-the-wild) deployment of such conversational agents, improving the Natural Language Understanding (NLU) module of the goal-oriented SDS pipeline is crucial, especially with limited task-specific datasets. This study explores the potential benefits of a recently proposed transformer-based multi-task NLU architecture, mainly to perform Intent Recognition on small-size domain-specific educational game datasets. The evaluation datasets were collected from children practicing basic math concepts via play-based interactions in game-based learning settings. We investigate the NLU performances on the initial proof-of-concept game datasets versus the real-world deployment datasets and observe anticipated performance drops in-the-wild. We have shown that compared to the more straightforward baseline approaches, Dual Intent and Entity Transformer (DIET) architecture~\cite{DIET-2020} is robust enough to handle real-world data to a large extent for the Intent Recognition task on these domain-specific in-the-wild game datasets.
 \\ \newline  }

\begin{document}

\maketitleabstract

\section{Introduction}

Investigating Artificial Intelligence (AI) systems that can help children in their learning process has been a challenging yet exciting area of research~\cite{CHASSIGNOL201816,zhai2021review}. Utilizing Natural Language Processing (NLP) for building educational games and applications has gained popularity in the past decade~\cite{lende2016question,cahill-etal-2020-context}. Game-based learning systems can offer significant advantages in teaching fundamental math concepts interactively, especially for younger students~\cite{skene2022can}. These intelligent systems are often required to handle multimodal understanding of the kids and their surroundings in real-time. Spoken Dialogue Systems (SDS) are vital building blocks for efficient task-oriented communication with children in game-based learning settings. In this study, the application domain is a multimodal dialogue system for younger kids learning basic math concepts through gamified interactions. Such dialogue system technology needs to be constructed and modeled carefully to handle task-oriented game interactions between the children and a virtual character serving as a conversational agent. 

Building the Natural Language Understanding (NLU) module of a goal-oriented SDS for game-based interactions usually involves: (i) the definition of intents (and entities if needed); (ii) creation of game-specific and task-relevant datasets; (iii) annotation of the game data with domain-specific intents and entities; (iv) iterative training and evaluation of NLU models; (v) repeating this tedious process for every new or updated game usages. Improving the NLU performances of task-oriented SDS pipelines in low-data regimes is quite challenging. This study primarily explores the potential benefits of a recent transformer-based multi-task architecture proposed for joint Intent and Entity Recognition tasks, especially with limited game datasets. Utilizing that flexible architecture, we focus on increasing the performance of our NLU models trained on small-size task-specific game datasets. The main NLU task we aim to improve is the Intent Recognition from possible user/player utterances during gamified learning interactions. Given an input utterance, the goal of an Intent Recognition model is to predict the user's intent (e.g., what the player wants to accomplish within a game-based interaction).

This work investigates the Intent Recognition model performances on our early proof-of-concept (POC) educational game datasets created to bootstrap the SDS to be deployed later in the real world. We have shown that adopting the recently proposed lightweight Dual Intent and Entity Transformer (DIET) architecture~\cite{DIET-2020} along with the Conversational Representations from Transformers (ConveRT) embeddings~\cite{ConveRT-2020} is a promising approach for NLU. This method boosts the NLU performance results on our initial small-scale POC game datasets. After the exploratory validation studies were conducted in-the-lab, the final evaluation datasets were collected in-the-wild from students working on fundamental math concepts in a game-based learning space at school. We examine the Intent Recognition performances on these real-world deployment datasets and reveal highly expected performance degradations in-the-wild. Compared to the baseline approaches, we have shown that adopting a DIET classifier with pre-trained ConveRT representations still achieves improved NLU results on our evaluation datasets collected in-the-wild.

\section{Related Work}

The use of AI technologies to enhance students' learning experiences has gained increasing popularity, especially in the last decade~\cite{CHASSIGNOL201816,10.1145/3290605.3300534,MMHCI,baker2021artificial,zhai2021review,ZHANG2021100025}. Intelligent game-based learning systems~\cite{Lester_Ha_Lee_Mott_Rowe_Sabourin_2013,10.1007/978-3-030-78292-4_28} present significant benefits for practicing math concepts in smart spaces~\cite{10.3389/feduc.2019.00081,sungamifying}, specifically for early childhood education~\cite{skene2022can}. Adapting NLP techniques to build various educational applications has been an appealing area of research for quite some time~\cite{meurers2012natural,10.1007/978-3-319-19773-9_3,lende2016question,taghipour-ng-2016-neural,raamadhurai-etal-2019-curio,cahill-etal-2020-context,ghosh-etal-2020-exploratory}. To slightly narrow down on these applications, building conversational agents for the smart education has been widely studied in the community~\cite{graesser2004autotutor,10.5555/1614025.1614027,10.1007/978-1-84882-215-3_13,Roos1223692,alex254848,PalaMohdNash2019gl,10.1145/3313831.3376781}. Relatively few number of studies also exist specifically on recognizing goals or intents of players in educational games~\cite{min2016player,min2017multimodal,hooshyar2019systematic}. 

Since our ultimate goal is to build dialogue systems for interactive educational games, we have outlined the previous studies with applications of AI and NLP for education context until now (e.g., intelligent systems and conversational agents for play-based learning). Next, we will briefly summarize the dialogue system technologies and NLU approaches in a more generic context.

Dialogue systems are frequently categorized as either task-oriented or open-ended. The task-oriented dialogue systems are designed to fulfill specific tasks and handle goal-oriented conversations. The open-ended systems or chatbots, on the other hand, allow more generic conversations such as chit-chat~\cite{Jurafsky:2000:SLP:555733}. With the advancements of deep learning-based language technologies and increased availability of large datasets with computing power in the research community, the dialogue systems trained end-to-end produce promising results for both goal-oriented~\cite{DBLP:journals/corr/bordes2016learning} and open-ended~\cite{DBLP:journals/corr/DodgeGZBCMSW15} applications. Dialogue Managers (DM) of goal-oriented systems are often sequential decision-making models. The optimal policies can be learned via reinforcement learning from a high number of user interactions~\cite{shah2016interactive,dhingra-etal-2017-towards,liu2017e2e,su-etal-2017-sample,cuayahuitl2017simpleds}. Unfortunately, building such systems with limited user interactions is extremely challenging. Therefore, supervised learning approaches with modular SDS pipelines are still widely preferred when initial training data is limited, basically to bootstrap the goal-oriented conversational agents for further data collection~\cite{KidSpace-SemDial-2019}. Statistical and neural network-based dialogue system toolkits and frameworks~\cite{bocklisch2017rasa,ultes-etal-2017-pydial,burtsev-etal-2018-deeppavlov} are heavily used in the academic and industrial research communities for implicit dialogue context management.

The NLU module within SDS pipeline processes the user utterances as input and often predicts the user intents (along with entities of interest if necessary). LSTM networks~\cite{lstm-1997} and Bidirectional LSTMs~\cite{bi-lstm-1997} have been widely utilized for sequence learning tasks such as Intent Classification and Slot Filling~\cite{mesnil-2015,hakkani2016multi}. Joint training of Intent Recognition and Entity Extraction models have been explored recently~\cite{zhang-2016,Liu+2016,goo-etal-2018-slot,varghese2020bidirectional}. Several hierarchical multi-task architectures are proposed for these joint NLU approaches~\cite{h-zhou-2016,wen-2018,AMIE-CICLing-2019,vanzo-etal-2019-hierarchical}, few of them in multimodal context~\cite{gu2017speech,okur-etal-2020-audio}. \newcite{DBLP:conf/nips/VaswaniSPUJGKP17} proposed the Transformer as a novel neural network architecture based entirely on attention mechanisms~\cite{DBLP:journals/corr/BahdanauCB14}.
Shortly after, Bidirectional Encoder Representations from Transformers (BERT)~\cite{devlin-etal-2019-bert} became one of the significant breakthroughs in pre-trained language representations, showing strong performance in numerous NLP tasks, including the NLU. Recently,~\newcite{DIET-2020} introduced the Dual Intent and Entity Transformer (DIET) as a lightweight multi-task architecture that outperforms fine-tuning BERT for predicting intents and entities on a complex multi-domain NLU-Benchmark dataset~\cite{Liu2021}. On the efficient representation learning side,~\newcite{ConveRT-2020} lately proposed the Conversational Representations from Transformers (ConveRT), which is also a lightweight approach to obtain pre-trained embeddings as sentence representations to be successfully utilized in numerous conversational AI tasks.

\begin{figure*}[t]
\begin{center}
\includegraphics[scale=0.31]{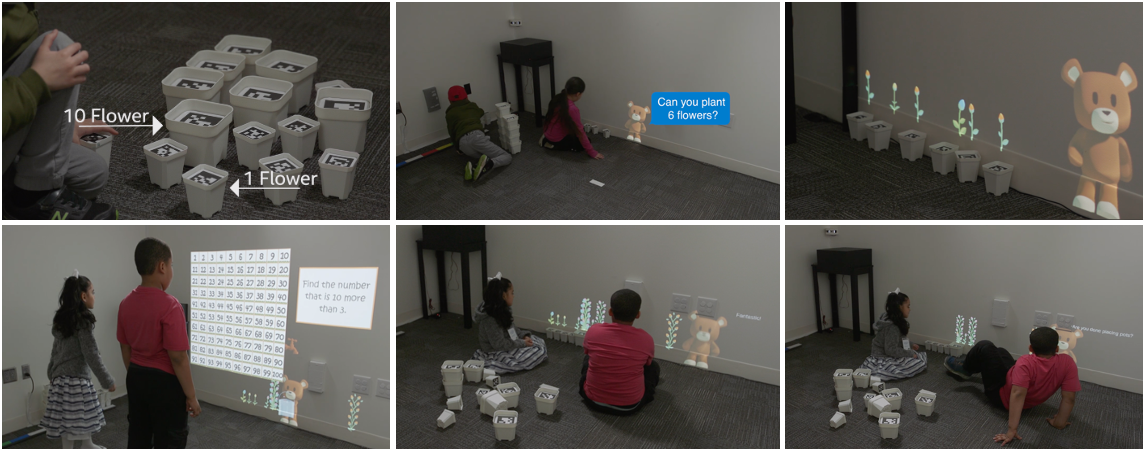} 
\caption{Learning basic math via game play-based interactions.}
\label{fig1-use}
\end{center}
\end{figure*}

\section{NLU Models}
\label{nlu-models}

This section describes the models we examine for the NLU (i.e., Intent Recognition) module within a dialogue system pipeline. We have built our NLU models on top of the Rasa open-source framework~\cite{bocklisch2017rasa}. The former baseline Intent Recognition architecture available in Rasa is based on supervised embeddings provided within the Rasa NLU~\cite{bocklisch2017rasa}, which is an embedding-based text classifier that embeds user utterances and intent labels into the same vector space. This former baseline architecture is inspired by the StarSpace work~\cite{wu2018starspace}, where the supervised embeddings are trained by maximizing the similarity between intents and utterances. The algorithm learns to represent user inputs and intents into a common embedding space and compares them against each other in that vectorial space. It also learns to rank a set of intents given a user utterance and provides similarity rankings of these labels. In~\newcite{KidSpace-SemDial-2019}, the authors enriched this embedding-based former baseline Rasa Intent Classifier by incorporating additional features and adapting alternative network architectures. To be more precise, they adapted the Transformer network~\cite{DBLP:conf/nips/VaswaniSPUJGKP17} and incorporated pre-trained BERT embeddings using the \texttt{bert-base-uncased} model~\cite{devlin-etal-2019-bert} to improve the Intent Recognition performance. In this work, we employed this improved approach from~\newcite{KidSpace-SemDial-2019} as our baseline NLU model, which we would call TF+BERT in our experiments.

In this study, we explore the potential improvements in Intent Classification performance by adapting the recent DIET architecture~\cite{DIET-2020}. DIET is a transformer-based multi-task architecture for joint Intent Recognition and Entity Extraction. It employs a 2-layer transformer shared for both of these NLU tasks. To be more precise, a sequence of entity labels is predicted with a Conditional Random Field (CRF)~\cite{DBLP:conf/icml/LaffertyMP01} tagging layer, which is on top of the transformer output sequence corresponding to the input sentences treated as a sequence of tokens. For the intent labels, the transformer output for the \texttt{\char`_\char`_CLS\char`_\char`_} token (i.e., classification token at the end of each sentence) and the intent labels are embedded into the same semantic vector space. The dot-product loss is utilized to maximize the similarity with the target label and minimize similarities with the negative samples. Note that DIET can incorporate pre-trained word and sentence embeddings from language models as dense features, with the flexibility to combine these with token level one-hot encodings and multi-hot encodings of character n-grams as sparse features. These sparse features are passed through a fully-connected layer with shared weights across all sequence steps. The output of this fully-connected layer is concatenated with the dense features from the pre-trained models. This flexible architecture allows us to use any pre-trained embeddings as dense features in DIET, such as GloVe~\cite{pennington-etal-2014-glove}, BERT~\cite{devlin-etal-2019-bert}, and ConveRT~\cite{ConveRT-2020}.

Conversational Representations from Transformers (ConveRT) is yet another recent and promising architecture to obtain pre-trained representations that are well-suited for real-world Conversational AI applications, especially for the Intent Classification task. ConveRT is a unique transformer-based dual-encoder network leveraging quantization and subword-level parameterization. In~\newcite{ConveRT-2020}, the authors show that pre-trained representations from the ConveRT sentence encoder can be transferred to the Intent Classification task with promising results. Both DIET and ConveRT are lightweight architectures with faster and memory/energy-efficient training capabilities than their counterparts. When incorporating the ConveRT embeddings with the DIET classifier, the initial embedding for \texttt{\char`_\char`_CLS\char`_\char`_} token is set as the input sentence encoding obtained from the ConveRT model. This way, we can leverage extra contextual information from the complete sentence on top of the word embeddings. For all the above reasons, we adapted the DIET architecture and incorporated pre-trained ConveRT embeddings to potentially improve the Intent Classification performances on our small domain-specific datasets. We would call this approach DIET+ConveRT in our experiments\footnote{Please refer to~\newcite{DIET-2020} for hyper-parameters, hardware specifications, and computational cost details.}. 

To investigate the actual benefits of DIET architecture versus the dense features, we also adapted DIET with out-of-the-box pre-trained BERT embeddings using the \texttt{bert-base-uncased} model~\cite{devlin-etal-2019-bert}, as in our baseline TF+BERT NLU model. When combining these off-the-shelf BERT representations with the DIET classifier, the initial embedding for \texttt{\char`_\char`_CLS\char`_\char`_} token is set to the corresponding output embedding of the BERT \texttt{[CLS]} token. We would call this approach DIET+BERT in our experiments.

\begin{table}[!t]
  \centering
  \resizebox{\columnwidth}{!}{
  \begin{tabular}{lcc}
    \toprule
     & \textbf{Planting} & \textbf{Watering} \\
    \textbf{Statistics/Dataset} & \textbf{Game} & \textbf{Game} \\
    \midrule
    \# distinct intents & 14 & 13 \\
    total \# samples (utterances) & 1927 & 2115 \\
    min \# samples per intent & 22 & 25 \\
    max \# samples per intent & 555 & 601 \\
    avg \# samples per intent & 137.6 & 162.7 \\
    \# unique words (vocab) & 1314 & 1267 \\
    total \# words & 10141 & 10469 \\
    min \# words per sample & 1 & 1 \\
    max \# words per sample & 74 & 65 \\
    avg \# words per sample & 5.26 & 4.95 \\
    \bottomrule
  \end{tabular}
  }
  \caption{KidSpace-PW-POC Dataset Statistics}
  \label{data-st-poc}
\end{table}

\section{Experimental Results}

\subsection{Datasets}
\label{datasets}

We conduct our experiments on the Kid Space Planting and Watering (PW) games NLU datasets having utterances from play-based math learning experiences designed for early school-age children (i.e., 5-to-8 years old)~\cite{KidSpace-ICMI-2018,KidSpace-ETRD-2022}. 
The use-cases aim to create an interactive smart space for children with traditional gaming motivations such as level achievements and virtually collecting objects. The smart space allows multiple children to interact, which can encourage social development. The intelligent agent should accurately comprehend inputs from children and provide feedback. The AI system needs to be physically grounded to allow children to bring meaningful objects into the play experience, such as physical toys and manipulatives as learning materials. Therefore, the multimodal system would combine various sensing technologies that should interact with children, track each child, and monitor their progress.

The use-cases include a specific flow of interactive games facilitating elementary math learning. The FlowerPot game (i.e., Planting Game in Tables~\ref{data-st-poc} and~\ref{data-st-dep}) builds on the math concepts of tens and ones, with the larger flower pots representing `tens' and smaller pots `ones'. The virtual character provides the number of flowers the children should plant, and when the children have placed the correct number of large and small pots against the wall, digital flowers appear. In the NumberGrid game (i.e., Watering Game in Tables~\ref{data-st-poc} and~\ref{data-st-dep}), math clues (or questions) are presented to children. When the correct number is touched on the number grid (i.e., on the wall), water is virtually poured to water the flowers. The visual, audio, and LiDAR-based recognition technologies enable physically situated interactions. The dialogue system is expected to take multimodal information to incorporate user identity, actions, gestures, audio context, and the objects (i.e., physical manipulatives) in space. For instance, during the FlowerPot game experience, the virtual character asks the children if they are done placing pots, to which they respond `yes' (or `no'). The dialogue system needs to use the visual input to have the virtual character respond appropriately to the correct (or incorrect) number of pots being detected. 

\begin{table}[!t]
  \centering
  \resizebox{\columnwidth}{!}{
  \begin{tabular}{lcc}
    \toprule
     & \textbf{Planting} & \textbf{Watering} \\
    \textbf{Statistics/Dataset} & \textbf{Game} & \textbf{Game} \\
    \midrule
    \# distinct intents & 12 & 11 \\
    total \# samples (utterances) & 2173 & 2122 \\
    min \# samples per intent & 4 & 6 \\
    max \# samples per intent & 1005 & 1005 \\
    avg \# samples per intent & 181.1 & 192.9 \\
    \# unique words (vocab) & 772 & 743 \\
    total \# words & 10433 & 9508 \\
    min \# words per sample & 1 & 1 \\
    max \# words per sample & 45 & 44 \\
    avg \# words per sample & 4.80 & 4.48 \\
    \bottomrule
  \end{tabular}
  }
  \caption{KidSpace-PW-Deployment Dataset Statistics}
  \label{data-st-dep}
\end{table}

Figure~\ref{fig1-use} demonstrates the virtual character (i.e., Oscar the teddy bear) helping the kids with learning `tens' and `ones' concepts along with practicing simple counting, addition, and subtraction operations. The game datasets have a limited number of player utterances, which are manually annotated for intent types defined for each learning game or activity. For the FlowerPot game, we use the Planting Flowers game dataset, and for the NumberGrid game, we use a separate Watering Flowers game dataset. Some of the intents are quite generic across usages and games/activities (e.g., \textit{affirm}, \textit{deny}, \textit{next-step}, \textit{out-of-scope}, \textit{goodbye}), whereas others are highly domain-dependent and game/task-specific (e.g., \textit{intro-meadow}, \textit{answer-flowers}, \textit{answer-water}, \textit{ask-number}, \textit{answer-valid}, \textit{answer-invalid}).

The current learning game activities are designed for two children collaboratively playing with the virtual agent. In addition to kids, an adult user (i.e., the Facilitator) is also present in the space to interact with the agent for game progress and help out the children when needed. Thus, we are dealing with a multiparty conversational system interacting with multiple users (i.e., two kids and one adult) while they progress through several learning games. In this goal-oriented dialogue system, the agent should provide the game instructions (with the Facilitator's help), guide the kids, and understand both the kids' and the Facilitator's utterances and actions to respond to them appropriately.

The NLU models are trained and validated on the initial POC datasets~\cite{KS2.0-NAACL-2021} to bootstrap the agents for in-the-wild deployments. These POC game datasets were manually created based on the User Experience (UX) design studies to train the SDS models and then validated with the UX sessions in-the-lab with 5 kids (and one adult) going through play-based learning interactions. Table~\ref{data-st-poc} shows the statistics of these KidSpace-PW-POC NLU datasets. Planting game and Watering game POC datasets have 1927 and 2115 utterances, respectively.

\begin{table}[!t]
  \centering
  \resizebox{\columnwidth}{!}{
  \begin{tabular}{llcc}
    \toprule
    \multicolumn{2}{c}{\textbf{Planting Game Datasets}} & \multicolumn{2}{c}{\textbf{\# Utterances}} \\
    \midrule
    \textbf{Type} & \textbf{Intent} &\textbf{POC} & \textbf{Deployment} \\
    \midrule
    \textbf{Domain} & \textit{intro-meadow} & 23 & 7 \\
    \textbf{Specific} & \textit{answer-flowers} & 110 & 13 \\
    & \textit{answer-valid} & 176 & 17 \\
    & \textit{answer-invalid} & 95 & 0 \\
    & \textit{intro-game} & 134 & 78 \\
    & \textit{help-affirm} & 41 & 4 \\
    & \textit{everyone-understand} & 22 & 11 \\
    & \textit{oscar-understand} & 25 & 15 \\
    & \textit{ask-number} & 34 & 18 \\
    & \textit{counting} & 418 & 581 \\
    \midrule
    \textbf{Generic} & \textit{affirm} & 144 & 370 \\
    & \textit{deny} & 125 & 54 \\
    & \textit{next-step} & 25 & 0 \\
    & \textit{out-of-scope} & 555 & 1005 \\
    \midrule
    & \textbf{Total} & 1927 & 2173 \\
    \bottomrule
  \end{tabular}
  }
  \caption{Intent Class Distributions for Planting Game}
  \label{data-dist-planting}
\end{table}

The deployment game datasets were later collected from 12 kids (and two adults), where the Kid Space setup was deployed in a classroom at school~\cite{KidSpace-ETRD-2022}. Table~\ref{data-st-dep} shows the statistics of KidSpace-PW-Deployment NLU datasets, where Planting game and Watering game deployment datasets have 2173 and 2122 utterances, respectively. Note that these deployment datasets are used only for the testing purposes in this study, where we train our NLU models on the POC datasets. For both in-the-lab and in-the-wild datasets, the spoken user utterances are transcribed manually at first. These transcriptions are then manually annotated for the intent types we defined for each game activity. These transcribed and annotated final utterances are analyzed and used in our experiments in this study.

When we compare the POC versus deployment game datasets (in Tables~\ref{data-st-poc}-to-~\ref{data-dist-watering}), we observe above 2.1k sample utterances for each game activity in both cases, except for the Planting POC data with around 1.9k samples. The number of possible user intents we envisioned for the POC was 14 and 13, respectively, for the Planting and Watering games. However, we have not observed any samples for two of the possible intent types for each game in the real-world deployment sessions. These intent types are \textit{next-step} and \textit{answer-invalid}, which were part of our backup intents in case we have technical issues and the users need to skip certain sub-activities (i.e., \textit{next-step}), or in case the users provide highly irrelevant or unexpected answers to our specific questions in the game flow (i.e., \textit{answer-invalid}). The minimum and the maximum number of samples per intent also differ significantly for the POC versus in-the-wild game datasets, which creates a huge difference in class distributions for our test samples (see Tables~\ref{data-dist-planting} and ~\ref{data-dist-watering}). Although we expect certain intent types to occur very infrequently in real game-plays (e.g., \textit{help-affirm}), we still have to manually create enough samples ($\geq$20) for each intent type for the model training and validation purposes during the POC. The dominant intent class in both POC and in-the-wild datasets is \textit{out-of-scope} (OOS). That was more or less anticipated as we are dealing with a multiparty conversational game setting here. In these games, the kids are encouraged to talk to each other while collaboratively solving the math puzzles. They can also discuss with or ask for help from the Facilitator. As the agent is in always-listening mode, if the users are not directly addressing Oscar, the system can detect those utterances as OOS (or \textit{counting}, which is the second most frequent intent class, depending on the context). Notice that POC datasets were created with around one-fourth of the utterances as OOS, whereas the deployment datasets have almost half of the utterances tagged as OOS. That was mainly due to a relatively talkative Facilitator at school and some kids' preferences to talk to the Facilitator more often than Oscar in real deployment sessions. We have observed this behavior less often in our in-the-lab UX sessions, as the adult in the room was one of the researchers guiding kids to talk to Oscar instead. Those out-of-distribution and unseen OOS samples create additional challenges for the NLU models when tested on in-the-wild game datasets. We have also observed the vocabulary sizes shrink in-the-wild as we tried to manually curate more variations in the POC datasets to make the NLU models more robust. The average number of tokens per sample (i.e., utterance length) is around 5 in the POC data, yet, we observe slightly shorter utterances in-the-wild that might affect the available contextual information.

\begin{table}[!t]
  \centering
  \resizebox{\columnwidth}{!}{
  \begin{tabular}{llcc}
    \toprule
    \multicolumn{2}{c}{\textbf{Watering Game Datasets}} & \multicolumn{2}{c}{\textbf{\# Utterances}} \\
    \midrule
    \textbf{Type} & \textbf{Intent} &\textbf{POC} & \textbf{Deployment} \\
    \midrule
    \textbf{Domain} & \textit{answer-water} & 69 & 9 \\
    \textbf{Specific} & \textit{answer-valid} & 201 & 6 \\
    & \textit{answer-invalid} & 91 & 0 \\
    & \textit{intro-game} & 102 & 30 \\
    & \textit{everyone-understand} & 44 & 11 \\
    & \textit{oscar-understand} & 25 & 15 \\
    & \textit{ask-number} & 73 & 21 \\
    & \textit{counting} & 476 & 581 \\
    \midrule
    \textbf{Generic} & \textit{affirm} & 165 & 370 \\
    & \textit{deny} & 157 & 54 \\
    & \textit{next-step} & 34 & 0 \\
    & \textit{out-of-scope} & 601 & 1005 \\
    & \textit{goodbye} & 77 & 20 \\
    \midrule
    & \textbf{Total} & 2115 & 2122 \\
    \bottomrule
  \end{tabular}
  }
  \caption{Intent Class Distributions for Watering Game}
  \label{data-dist-watering}
\end{table}

\subsection{Intent Recognition Results}

To evaluate the Intent Recognition performances, the baseline NLU model that we previously explored, TF+BERT, is compared with the DIET+BERT and DIET+ConveRT models that we adapted recently (see section~\ref{nlu-models}). We conduct our evaluations on both the Planting and Watering game datasets. The models are trained and validated on the bootstrap POC datasets and then tested on the school deployment datasets. Table~\ref{nlu-results-poc} summarizes the Intent Classification performance results on the POC datasets in micro-average F1-scores. To test our model extensively on these limited-size POC datasets, we perform a 10-fold cross-validation (CV) by automatically creating multiple train/test splits. We report the average performance results with standard deviations obtained from the 3 runs, where we perform a 10-fold CV over the POC datasets for each run. As one can observe from Table~\ref{nlu-results-poc}, adapting the lightweight DIET architecture~\cite{DIET-2020} with pre-trained ConveRT embeddings~\cite{ConveRT-2020} significantly improved the Intent Classification performances for the NLU datasets manually created for POC. Specifically, the overall NLU performance gains are higher than 5\% F1-scores for both Planting and Watering game datasets. Note that when we keep the dense features (i.e., pre-trained embeddings from BERT language models) constant, we can observe the clear benefits of switching from standard Transformer (TF) architecture to DIET classifier. We gain 3-to-4\% F1-scores with DIET architecture, and we improve the Intent Recognition performance by another 1-to-2\% with ConveRT embeddings compared to BERT. With these observations, which are consistent across different use-cases (i.e., Planting and Watering games), we updated the NLU component in our multimodal SDS pipeline by replacing the previous TF+BERT model with this promising DIET+ConveRT architecture.

\begin{table}[!t]
  \centering
  \begin{tabular}{lcc}
    \toprule
     & \textbf{Planting} & \textbf{Watering} \\
    \textbf{Model/Dataset} & \textbf{Game} & \textbf{Game} \\
    \midrule
    TF+BERT (Baseline) & 90.50$\pm$0.25 & 92.43$\pm$0.32 \\
    \midrule
    DIET+BERT & 94.00$\pm$0.38 & 96.39$\pm$0.14 \\
    DIET+ConveRT & \textbf{95.88}$\pm$\textbf{0.42} & \textbf{97.69}$\pm$\textbf{0.11} \\
    \midrule
    Performance Gain & +5.38 & +5.26 \\
    \bottomrule
  \end{tabular}
  \caption{NLU/Intent Recognition micro-avg F1-scores (\%): TF+BERT, DIET+BERT, and DIET+ConveRT models trained and validated on KidSpace-PW-POC datasets (3 runs of 10-fold CV)}
  \label{nlu-results-poc}
\end{table}

Next, we investigate the NLU model performances on our real-world deployment datasets. The anticipated performance drops occurred when we tested these Intent Recognition models on in-the-wild data, which reflect more realistic game settings from a school deployment. Table~\ref{nlu-results-dep} summarizes the Intent Classification performance results obtained on the deployment game datasets in micro-average F1-scores. Although the DIET+ConveRT models trained on POC datasets performed very well during the cross-validation (i.e., achieved around 96\% and 98\% F1-scores for Planting and Watering games, respectively), the performance loss is significantly high (i.e., around 7\% F1-score) when tested on in-the-wild datasets. As a result, the same models achieved around 89\% and 91\% F1-scores when tested on the Planting and Watering deployment sets, respectively. That finding is quite common and probably not very surprising as the players in-the-wild can often largely deviate from the manual or synthetic data generation inside the labs or data collection through crowd-sourcing for interactive games. We have summarized the game dataset statistics and our preliminary observations regarding the main differences between the POC and deployment sets in section~\ref{datasets}. We believe such deviations have played a significant role in the observed performance shifts for real-world play-based interactions. More specifically, the sample-class distributions, vocabulary sizes, slightly shorter utterance lengths, frequency of the OOS conversations due to multiparty setup, technical issues during the sessions causing unexpected interactions, etc., would all contribute to these shifts. One should also keep in mind the unprecedented group dynamics for that age group in play-based interactions and the unpredictable nature of kids in game-based learning settings. These factors also play some role in the robustness issues of NLU models developed for such challenging real-world deployments.

\begin{table}[!t]
  \centering
  \begin{tabular}{lcc}
    \toprule
     & \textbf{Planting} & \textbf{Watering} \\
    \textbf{Model/Dataset} & \textbf{Game} & \textbf{Game} \\
    \midrule
    TF+BERT (Baseline) & 85.08$\pm$0.49 & 90.06$\pm$0.56 \\
    \midrule
    DIET+BERT & 87.03$\pm$0.30 & 89.63$\pm$0.62 \\
    DIET+ConveRT & \textbf{89.00}$\pm$\textbf{0.29} & \textbf{90.57}$\pm$\textbf{0.86} \\
    \midrule
    Performance Gain & +3.92 & +0.51 \\
    \bottomrule
  \end{tabular}
  \caption{NLU/Intent Recognition micro-avg F1-scores (\%): TF+BERT, DIET+BERT, and DIET+ConveRT models trained on KidSpace-PW-POC (3 runs) and tested on KidSpace-PW-Deployment datasets (3 runs)}
  \label{nlu-results-dep}
\end{table}

\begin{table*}[t!]
  \centering
  \footnotesize
  \resizebox{\textwidth}{!}{
  \begin{tabular}{clcc}
    \toprule
    \textbf{Data} & \textbf{Sample Utterance} & \textbf{Intent} & \textbf{Prediction} \\
    \midrule
    \textbf{Planting} & oh so like green and blue colors? & \textit{answer-valid} & \textit{answer-flowers} \\
    \textbf{Game} & thirteen flowers! & \textit{counting} & \textit{answer-flowers} \\
    & so if we had to start at a number what number do you think we should start at? & \textit{counting} & \textit{ask-number} \\
    & or twenty less okay so we're going down & \textit{counting} & \textit{out-of-scope} \\
    & let's add let's add a flower pot what do you think? & \textit{counting} & \textit{intro-game} \\
    & yeah totally! do you wanna plant some next to him? & \textit{affirm} & \textit{intro-game }\\
    & yeah I think that's ninety & \textit{affirm} & \textit{counting} \\
    & no I think it was forty five & \textit{deny} & \textit{counting} \\
    & okay so what do we need to start with? & \textit{out-of-scope} & \textit{help-affirm} \\
    \midrule
    \textbf{Watering} & next one? & \textit{ask-number} & \textit{next-step} \\
    \textbf{Game} & okay so how many more do you think we need? & \textit{counting} & \textit{ask-number} \\
    & we need ten more to water & \textit{counting} & \textit{answer-water} \\
    & to give to have enough water to plant our flowers and make them grow & \textit{intro-game} & \textit{answer-water} \\
    & so when we look at these numbers all of the ones with the two in front, have two tens & \textit{intro-game} & \textit{counting }\\
    & if we get four correct answer & \textit{intro-game} & \textit{counting} \\
    & all right he's gotta go get his watering can that he must have put it away & \textit{out-of-scope} & \textit{intro-game} \\
    & the ground & \textit{out-of-scope} & \textit{answer-valid} \\
    & timber what & \textit{out-of-scope} & \textit{answer-valid} \\
    \bottomrule
  \end{tabular}
  }
  \caption{NLU/Intent prediction error samples from Planting and Watering games deployed in-the-wild: DIET+ConveRT model trained on KidSpace-PW-POC datasets and tested on KidSpace-PW-Deployment datasets}
  \label{errors}
\end{table*}

Besides the inevitable NLU performance degradations on real-world deployment datasets, Table~\ref{nlu-results-dep} also compares the baseline TF+BERT models with more recent DIET+BERT and DIET+ConveRT architectures, all trained on POC data and tested on in-the-wild game data. The DIET+ConveRT models still reach the highest Intent Recognition F1-scores on these test sets, but the gap between the baseline and the best-performing models has been narrowed, especially for the Watering game. Compared to the TF+BERT baseline, the performance gain with the DIET+ConveRT approach is +3.92\% in Planting and only +0.51\% in Watering games when tested in-the-wild. For Planting, the increasing performance trends going from TF+BERT to DIET+BERT and DIET+ConveRT are also distinguishable on the deployment set. However, for Watering, the baseline TF+BERT model performs relatively well when tested on the deployment set, achieving only slightly lower F1-scores than the DIET+ConveRT. Notice that the variances are also relatively high in this case, so we may not observe the significant performance benefits when switching to DIET architecture from baseline TF for Watering game deployment. The possible reason for the baseline model in Watering being already quite robust on real-world data could be the size differences in POC datasets on which the models are trained. To be more precise, the Planting baseline model is trained on 1927 samples and tested on 2173 in-the-wild utterances (see Table~\ref{data-dist-planting}). Unlikely, the Watering baseline model is trained on 2115 samples and tested on 2122 utterances (see Table~\ref{data-dist-watering}). In addition, we have one less intent class to predict in total (e.g., 14 vs. 13) and two fewer domain-specific intent types (e.g., 10 vs. 8) in the Watering game compared to the Planting. Having around 10\% more data for training, plus having slightly less number of total and domain-specific intents, can explain the relatively higher robustness of the baseline model on Watering deployment data (compared to Planting). On the other hand, due to the consistently significant improvements obtained in all other cases (i.e., Planting-POC, Watering-POC, and Planting-Deployment), DIET+ConverRT still seems a promisingly more robust NLU model for our future use-cases.

\subsection{Error Analysis and Discussion}

In this subsection, we aim to investigate further the differences between the POC and the real-world deployment datasets for NLU in our game-based learning activities. When best-performing DIET+ConveRT models were tested in-the-wild, we discovered overall F1-score performance drops of around 7\% for Intent Recognition, consistently for both game activities (i.e., Planting and Watering). When we analyze the intent-wise results, we identify some generalization issues between the POC to in-the-wild datasets, especially with the highly domain-specific intents. 

For the Planting Flowers game, the top 5 intent classes with highest performance drops ($\geq$20\%) are \textit{answer-valid}, \textit{help-affirm}, \textit{ask-number}, \textit{answer-flowers}, and \textit{intro-game}. Among these, \textit{help-affirm} had quite low test samples (only 4 utterances observed during deployments), which could explain the high variance in the detection performance. Regarding these top 5 erroneous intent classes, we realize that these are highly domain-dependent and activity-specific intent types, where we expect vastly specific answers from the kids based on the game flow design. To illustrate, during this Planting game, kids are helping Oscar to make the meadow look nicer. At the beginning of their interactions, the virtual character is asking ``\textit{Let's see, what could we add to the meadow... What kind of plants have pretty colors and smell nice?}" (or something along those lines as we use variations in response templates). We expect the kids to answer with ``\textit{flowers}" or its variations at this point in the game, where these short utterances should be classified as \textit{answer-flowers} intent. However, kids can also answer with other plants (or animals, etc.) that belong to the meadow, like ``\textit{trees}", ``\textit{bushes}", ``\textit{butterflies}", ``\textit{birds}", etc. These viable but incorrect answers would ideally be classified as \textit{answer-valid} intent. As you can see, these are extremely task-specific intents, and numerous things can go wrong in-the-wild for these, which may be beyond our assumptions. The \textit{intro-game} intent is also highly game-specific as it is designed to cover the possible utterances from the Facilitator while s/he is introducing the game and explaining the rules (e.g., how to use the big and small pots for `tens' and `ones' for this Planting Flowers game). For more generic intent types that can be shared across other activities (e.g., \textit{affirm}, \textit{deny}, \textit{out-of-scope}), we observed relatively less performance degradation in-the-wild using DIET+ConveRT.

For the Watering game activity, the top 4 intents with highest performance degradations ($\geq$20\%) are \textit{answer-valid}, \textit{ask-number}, \textit{intro-game}, and \textit{answer-water}. This time, \textit{answer-valid} had very few test samples (only 6 utterances observed during the Watering game at school sessions), which might again explain the high variance in its performance. All these four intent types are also highly task-specific, and we anticipate more vulnerability for deviations in-the-wild for them, in contrast to the generic intent classes (e.g., \textit{affirm}, \textit{deny}, \textit{out-of-scope}, \textit{goodbye}). During the Watering game, this time, Oscar is asking ``\textit{What do you think we need to help the flowers bloom?}". We expect the kids to answer with ``\textit{water}" or its variations, where such utterances should be recognized as \textit{answer-water} intent. Once again, kids can say other viable answers that could help the flowers grow/bloom, such as ``\textit{sunlight}", ``\textit{soil}", ``\textit{bees}", etc., which should be classified as \textit{answer-valid} intent. Similarly, the \textit{intro-game} intent is extremely domain/game-specific and aims to detect Facilitator utterances while s/he is introducing/explaining the game rules (e.g., how to use the number grid projected on the wall for touch-based interactions in this Watering game). Note that these valid answers or game introductions differ substantially based on which game we are playing, and we need to train separate NLU models for each game using these game domain-specific samples.

Table~\ref{errors} depicts some of the user utterances collected in-the-wild as concrete examples from both deployment datasets. The ground truth intent labels and the predicted intent classes are compared, emphasizing the errors made on some of the most problematic game-specific intents. Here we use our best-performing DIET+ConveRT models for the Intent Classification task. These prediction errors are expected to occur in real-world deployments for various reasons. Some of these user utterances could have multiple intents (e.g., ``\textit{yeah totally! do you wanna plant some next to him?}" starts with \textit{affirm}, then the Facilitator continues guiding the kids during \textit{intro-game}). Others could fail due to subtle semantic differences between these classes (e.g., ``\textit{if we get four correct answer}" is used by the Facilitator while explaining the NumberGrid game rules but can easily be mixed with \textit{counting} too). There exist some utterances where we see ``\textit{flowers}" or ``\textit{water}" while \textit{counting} with numbers (e.g., ``\textit{thirteen flowers!}", ``\textit{we need ten more to water}"), which are confusing for the models trained on much cleaner datasets. Note that the majority of these classification errors occur for the user utterances during multiparty conversations, i.e., the users are talking to each other instead of Oscar, the virtual game character, but the SDS fails to recognize that (e.g., ``\textit{okay so what do we need to start with?}"). These sample utterances also depict several cases where our highly vocal adult Facilitator at school is talking to the kids to introduce the games, explain the rules, guide them to count loudly, and help them find the correct answers in the game flow. It is highly challenging to predict those nearly open-ended conversations and include all possibilities in the POC training datasets to make the NLU models more robust for real-world deployments. However, we are working towards clustering-based semi-supervised intent discovery and human-in-the-loop (HITL) bulk labeling approaches~\cite{KS2.0-NAACL-2021,shen-etal-2021-semi} for cleaner design and separation of intent classes on in-the-wild datasets. We also plan to continue our data augmentation with paraphrase generation efforts to increase the limited POC samples and add more variations during training to make the NLU models more robust in future deployments\cite{okur2022data}.

\section{Conclusion}

Dialogue systems are vital building blocks to carry out efficient task-oriented communication with children for game play-based learning settings. This study investigates a small step towards improving contextually aware multimodal agents that need to understand and track children's activities and interactions during educational games, support them in performing learning tasks and provide insights to teachers and parents to help personalize the learning experiences. We focus on building task-specific dialogue systems for younger kids learning basic math concepts via gamified interactions. We aim to improve the NLU module of the goal-oriented SDS pipeline with domain-specific game datasets having limited user/player utterances.

In this exploration, we experimented with a flexible and lightweight transformer-based multi-task architecture called DIET~\cite{DIET-2020} to improve the NLU performances on our task-specific game datasets with limited sizes. These domain-specific datasets are manually created for the POC first and then tested on in-the-wild deployment data. Based on the results obtained on POC game datasets, using the DIET classifier with pre-trained ConveRT embeddings has shown to be a promising approach yielding remarkably higher F1-scores for Intent Classification. The NLU results on the real-world deployment game datasets also support these preliminary findings but to a lesser extent.

Using the best performing DIET+ConveRT approach, we observed significant performance drops when the NLU models were tested on in-the-wild game datasets compared to the initial POC datasets. That finding was foreseeable as the player utterances in real-world deployments may usually diverge from the samples within the POC data manually generated for bootstrapping purposes. We investigated these game datasets and shared our exploratory insights for the deviations between POC and in-the-wild datasets. Our preliminary observations suggest that the highest performance shifts occur for the more domain-specific intents in each educational game set. We are working towards making the NLU models and eventually the SDS pipeline more robust for such deviations in-the-wild by empowering the interactive intent labeling with HITL learning techniques and the data augmentation with paraphrasing.

\section{Acknowledgements}

We thankfully show our gratitude to our current and former colleagues from the Intel Labs Kid Space team, especially Ankur Agrawal, Glen Anderson, Sinem Aslan, Benjamin Bair, Arturo Bringas Garcia, Rebecca Chierichetti, Hector Cordourier Maruri, Pete Denman, Lenitra Durham, Roddy Fuentes Alba, David Gonzalez Aguirre, Sai Prasad, Giuseppe Raffa, Sangita Sharma, and John Sherry, for the conceptualization and the design of use-cases to support this research. In addition, we gratefully acknowledge the Rasa team for the open-source framework and the community developers for their contributions that enabled us to improve our research and build proof-of-concept models for our use-cases.

\section{Bibliographical References}\label{reference}

\bibliographystyle{lrec2022-bib}
\bibliography{lrec2022-example}

\label{lr:ref}
\bibliographystylelanguageresource{lrec2022-bib}

\end{document}